\documentclass[letterpaper]{article}
\usepackage{iccc}
\usepackage{times}
\usepackage{helvet}
\usepackage{courier}
\usepackage{xcolor}
\usepackage{url}

\usepackage{graphicx}
\usepackage{amsmath,amssymb} 
\usepackage{color}
\usepackage{epsfig}
\usepackage{pdfpages}
\usepackage{caption}
\usepackage{subcaption}

\usepackage{multirow}
\usepackage{rotating}
\usepackage{booktabs}
\usepackage{algpseudocode}
\usepackage{algorithm}

\newenvironment{packed_itemize}{
\begin{list}{\labelitemi}{\leftmargin=1em}
\vspace{-4pt}
  \setlength{\itemsep}{0pt}
  \setlength{\parskip}{0pt}
  \setlength{\parsep}{0pt}
}{\end{list}}

\title{Predicting A Creator's Preferences In, and From, Interactive Generative Art}

\author{Devi Parikh\\
Georgia Tech \& Facebook AI Research\\
parikh@gatech.edu\\
}

\begin{document} 
\maketitle


\begin{abstract}
\begin{quote}
As a lay user creates an art piece using an interactive generative art tool, what, if anything, do the choices they make tell us about them and their preferences? These preferences could be in the specific generative art form (e.g., color palettes, density of the piece, thickness or curvatures of any lines in the piece); predicting them could lead to a smarter interactive tool. Or they could be preferences in other walks of life (e.g., music, fashion, food, interior design, paintings) or attributes of the person (e.g., personality type, gender, artistic inclinations); predicting them could lead to improved personalized recommendations for products or experiences.

To study this research question, we collect preferences from 311 subjects, both in a specific generative art form and in other walks of life. We analyze the preferences and train machine learning models to predict a subset of preferences from the remaining. We find that preferences in the generative art form we studied cannot predict preferences in other walks of life better than chance (and vice versa). However, preferences within the generative art form are reliably predictive of each other. 

\end{quote}
\end{abstract}

\section{Introduction}

Generative art is art that has at least some of its features determined by a non-human autonomous system.\footnote{Generative Art (Wikipedia): \url{https://en.wikipedia.org/wiki/Generative_art}} The autonomous system is typically a computer, and it frequently relies on randomization to determine the art features. Among the human-defined features, some might be fixed by a generative artist, and other parameters might be open to manipulation. Different settings of these open parameters, combined with the machine-defined features, leads to different instances of the generative art form. Generative artists often make interactive tools available so a lay person can set values of these open parameters and create their own generative art piece.  We refer to these tools as interactive generative art tools, and they are the subject of this study.

We ask the research question: what does the choices a lay person makes while creating art using an interactive generative art tool tell us about them -- about their personality or preferences in food, fashion, interior design, etc., as well as about their preferences in the specific generative art form? 

Effectively predicting their preferences in the specific generative art form can lead to a smarter interactive generative art tool. It can help the user create an art piece they like faster by encouraging them to explore a certain part of the parameter space. It can prevent the user from losing interest by discouraging them to explore a different part of the parameter space. Predicting their preferences in other aspects of life can position interactive art generation as a generic personality assessment tool for products and experiences recommendations. Finally, predicting their preferences in generative art from other other known preferences in life can lead to improved art recommendation.

We conducted a survey where 311 subjects consented to participate and self-reported their preferences along various parameters in a generative art form, as well as in various walks of life such as food, chocolate, alcoholic beverages, music, interior design, fashion, paintings and their other traits such as gender, personality type,  exposure to design principles, artistic inclination, and introspectiveness.

We analyze these preferences to identify statistical correlations between pairs of preferences and report some interesting findings. We train machine learning models to predict subsets of these preferences from other preferences. In addition to some black box models, we include models that reveal interpretable relationships between various preferences.

Quantitatively, we find that for the specific generative art and machine learning models we experimented with, user's preferences in other aspects of life cannot be reliably predicted from their preferences in the generative art form (and vice versa). However, their preferences in the generative art form can be predicted with statistical significance from other preferences in the generative art form. This is a promising result towards demonstrating the feasibility of smarter interactive generative art tools.

\begin{figure*}[t]
    \centering
    \includegraphics[width=\textwidth]{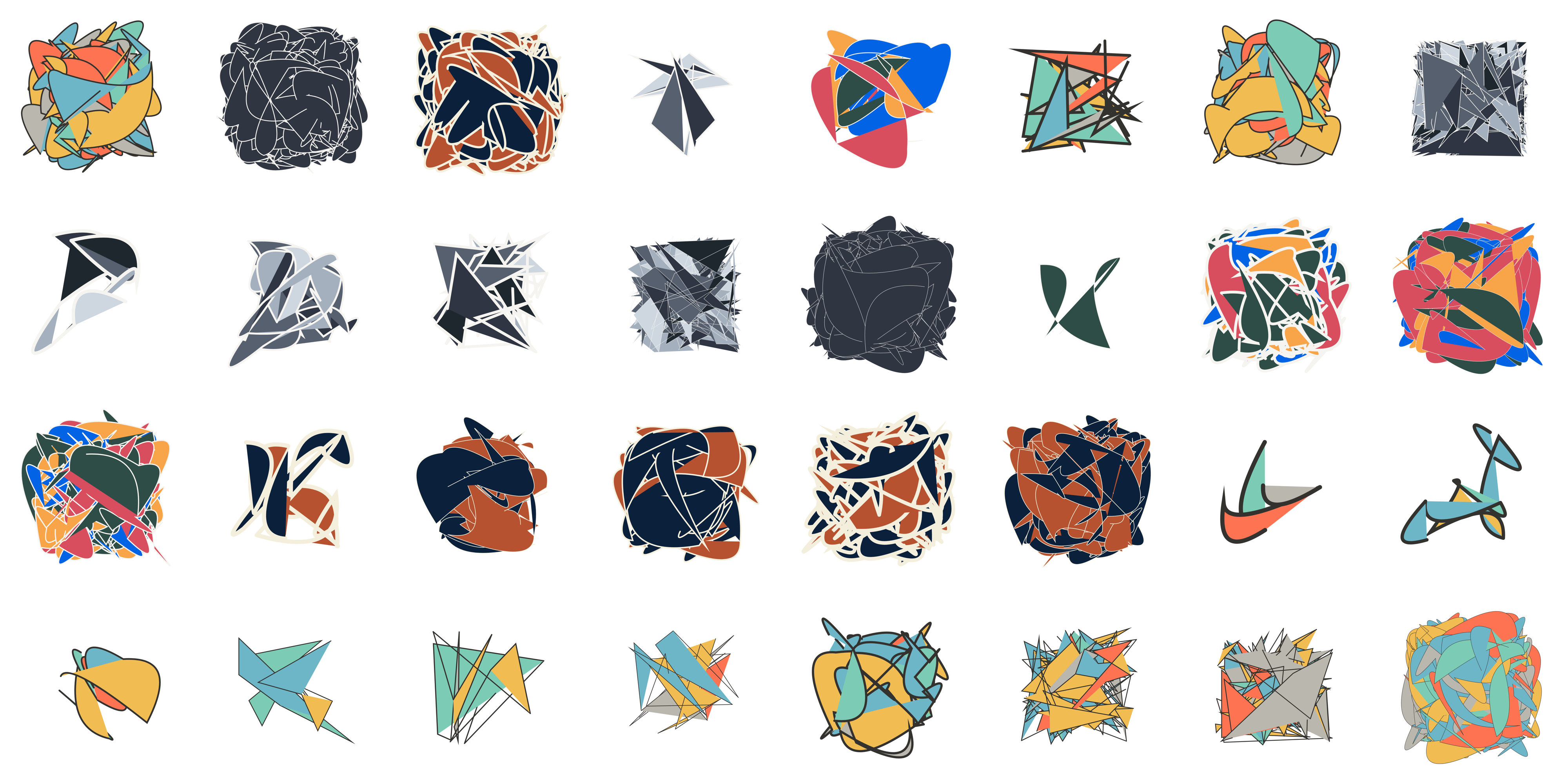}
    \caption{Example instances of the Strokes generative art form that we study in this work.}
    \label{fig:different_strokes}
\end{figure*}

\section{Related Work}

\begin{figure}[t]
    \centering
    \includegraphics[width=\columnwidth]{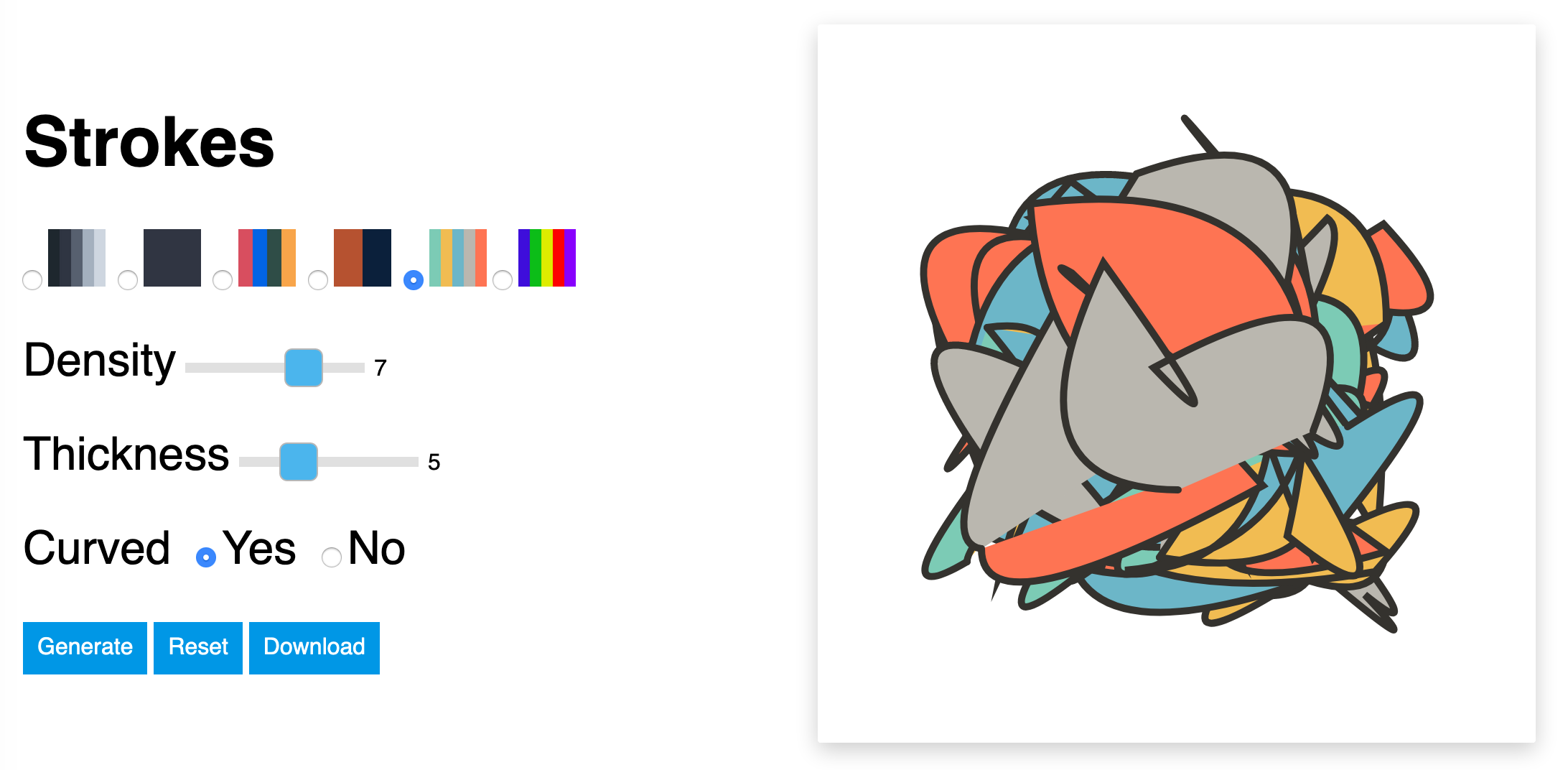}
    \caption{Interactive generative art tool for creating Strokes. Video: \url{https://youtu.be/YzfzjK8NNMg}.}
    \label{fig:interactive_strokes}
\end{figure}
\begin{figure*}[t]
    \centering
    \includegraphics[width=\textwidth]{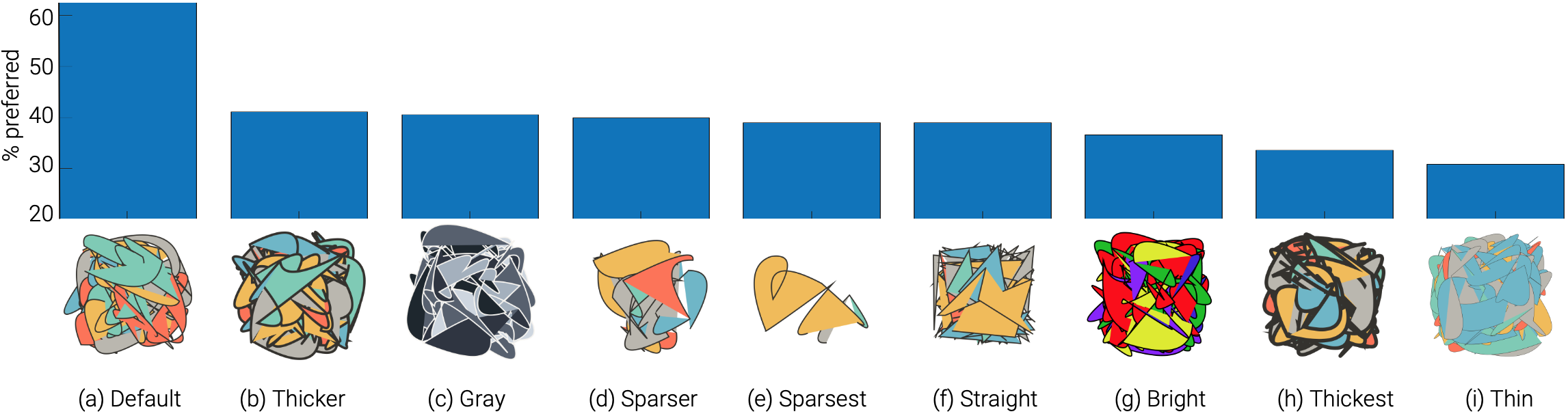}
    \caption{Configurations of Strokes we studied and \% times each alternative (b-i) was preferred by subjects over the default (a).}
    \label{fig:default_vs_alternatives}
\end{figure*}

\textbf{Preferences in art and personality.} Connections between perception of and preferences in art, and personality traits or preferences in other domains has been studied in several works. We describe those next. However, to the best of our knowledge, this has not been studied in the context of generative art specifically or even digital art in general. As the landscape of art changes with the incorporation of digital tools and more recently AI, it is valuable to consider how these tools can be made smarter to better assist a human in their creative process. The ability to predict a user's preferences is central to such smart tools.

\cite{Oz2015} study the correlations between five personality traits and art preferences among 24 visuals from Renaissance, Cubism, Abstract Art, Traditional Art, Impressionism and Surrealism. They found that extroversion and openness to experience was correlated with a preference for Surrealist works, while tender-mindedness was correlated with Impressionist works. \cite{Chamorro-Premuzic2008} performed a similar study across over 90,000 individuals in UK. \cite{Chamorro-Premuzic2010} find that personality traits correlate better with art categories when these categories are defined based on emotional valence and complexity as assessed by a collection of observers, than categories defined by researchers or historical art taxonomies. \cite{Reic2015} also study the relationship between personality traits and art preferences, but for both visual art and music. They also investigate correlations between music and visual art preferences. \cite{Gridley2013} study not only the correlations of visual art preferences with personality traits, but also with styles of thinking (concrete versus abstract thinking, sequential versus random thinking, ego functions of intuition and sensation). They also find evidence for cross-modal relations in aesthetic preferences across food, music, and visual stimuli. \cite{Lyssenko2015} study how participants describe abstract artwork and the relationship of these descriptions to various image properties. They also investigate the correlation between personality traits and preferences. They found neuroticism to correlate with a preference for objectively more complex images. 

\cite{Bhattacharjya2016} discusses formal models of preferences in the context of computational creativity to embrace the subjective nature of an evaluator judging creative value. \cite{Cook2015} design a software system capable of having preferences -- to make and justify subjective decisions beyond using random chance or a pre-defined external heuristic. Authors argue that having such preferences adds to the perception of the software being creative.

\textbf{Casual creators.} The interactive generative art tools we study fall in the category of ``Casual Creators''. This is in contrast with tools that are designed to assist task-oriented professionals or amateurs in their creative process towards solving a specific problem or designing for a specific task. Casual creators on the other hand are autotelic creativity tools that cater to enjoyable explorative creativity over task completion. \cite{Compton2015}, who coined the term, states  ``A Casual Creator is an interactive system that encourages the fast, confident, and pleasurable exploration of a possibility space, resulting in the creation or discovery of surprising new artifacts that bring feelings of pride, ownership, and creativity to the users that make them.'' 

They stress the interactive aspect of casual creators where the user is the driver, and the creating process as being core to the experience. A casual creator is an effective tool if it helps users find desirable artifacts without getting stuck in a local minima or being lost in a vast space of bad artifacts. Our work addresses exactly this. The various parameters in the interactive generative art tool define the space of artifacts, of possibilities, that a user can explore. Our work predicts the user's preferences. A computational model that uses these predictions can influence the path the user takes in this space; it affects the probability that a user will encounter a certain artifact in their creating process. By effectively predicting their preferences, we increase the chances that users will find a desirable artifact when using a casual creator.

\textbf{Individual user preferences.} Existing work, e.g.,~\cite{Zsolnai-Feher}, models \emph{individual} preferences of a user as they explore a parametrized design space. Our work learns correlations between preferences across parameters from a \emph{population} of users. The two directions have a common goal -- helping a user find designs they like -- but are complementary.

\textbf{GANs.} Generative Adversarial Networks~\cite{gans} have gained tremendous popularity as tools for AI-powered generative visual art. In this work we focus on a more geometric, abstract, flat, generative art form that gives the user direct control over specific features of the art. 

\section{Interactive Generative Art: Strokes}

We start by defining the generative art form --  Strokes -- around which we design our study. Example art pieces of the Strokes form are shown in Figure~\ref{fig:different_strokes}. We chose Strokes because it is an abstract form, allowing us to focus our study on visual preferences rather than semantic associations.

A Strokes piece is a series of overlaid shapes. A shape is started by connecting two random points on a square canvas via a curve of a certain thickness ($T$). The curve may be a straight line, or a quadratic Bezier curve using a third point as a control point. This control point is chosen to be the midpoint between the two end points perturbed by random noise. The random noise lies uniformly randomly in the range $R$, which is 10\% to 20\% the width of the canvas. This noise is either added or subtracted to the $x$ and $y$ co-ordinates of the mid-point independently. Each of these 4 possibilities has a probability $\nu$ of 0.25.

Having placed the first curve, the end point of the curve is connected to another random point on the canvas via a curve (again, straight line or quadratic Bezier curve with a noisy control point). This process is repeated. After each curve is drawn, the shape either continues (with probability $P = 0.5$) or the shape ends and a new shape begins. When the shape ends, the canvas enclosed by the curves and a straight line connecting the start point of the first curve in the shape and end point of the last curve in the shape is colored by a random color from a palette. The color of the curves themselves is a pre-defined background color in the palette. When a total of $N$ curves have been drawn, that shape and the piece overall are complete. Any pixel covered by more than one shape is colored by the most recent color.

\begin{figure*}[t]
\centering
\begin{subfigure}[b]{0.15\textwidth}
	\centering
	\includegraphics[height=0.5\textwidth]{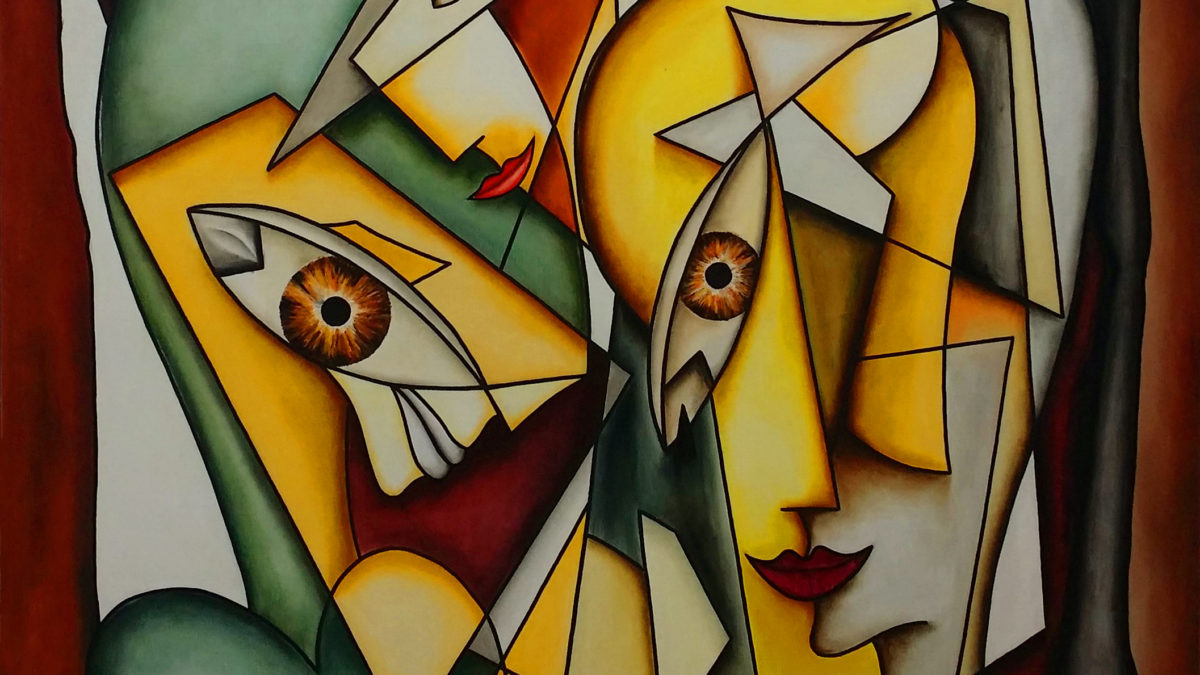} 
	\caption{Cubism}
	\label{fig:painting_l}
\end{subfigure}
\quad
\begin{subfigure}[b]{0.15\textwidth}
	\centering
	\includegraphics[height=0.5\textwidth]{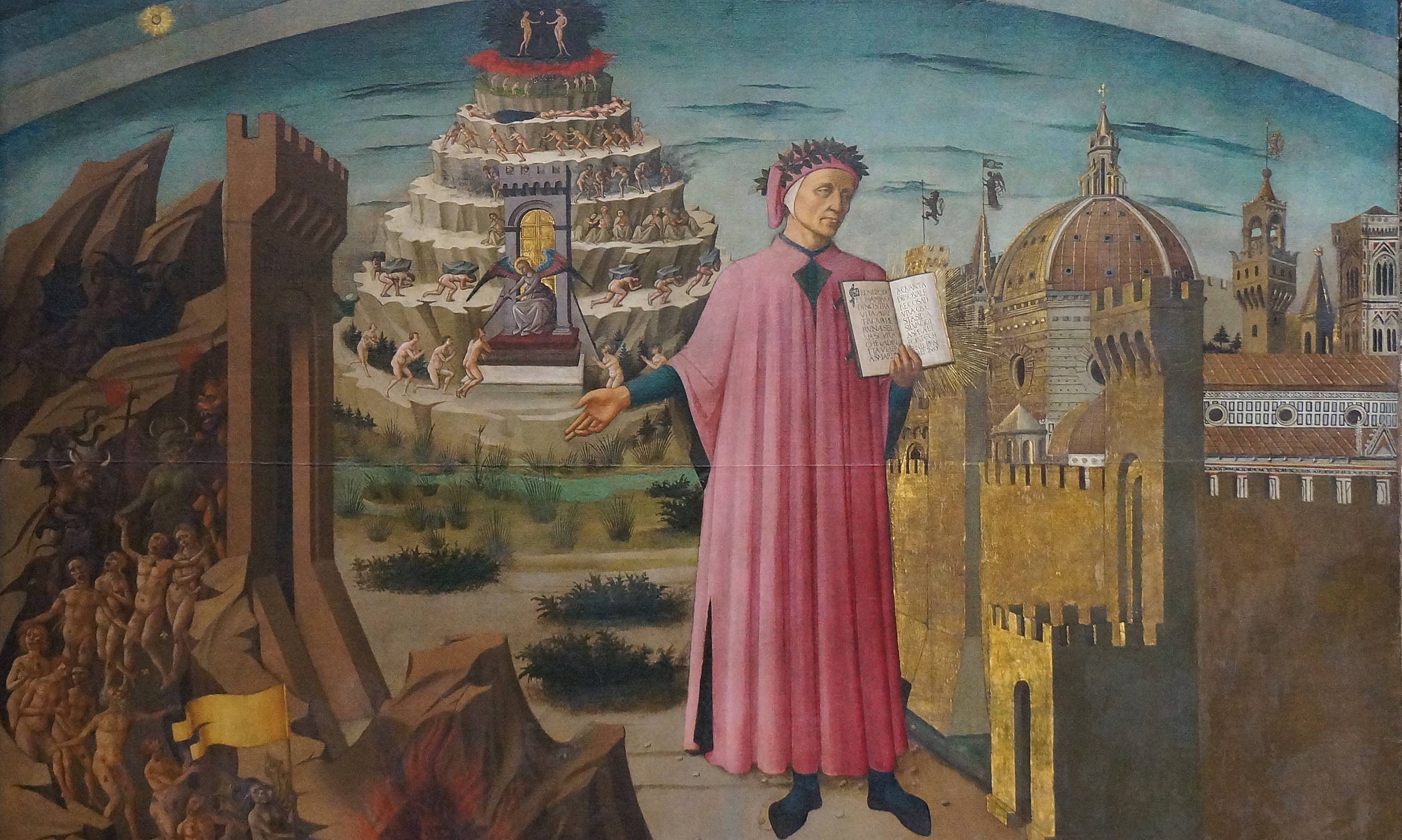} 
	\caption{Renaissance}
	\label{fig:painting_r}
\end{subfigure}
\begin{subfigure}[b]{0.15\textwidth}
	\centering
	\includegraphics[height=0.5\textwidth]{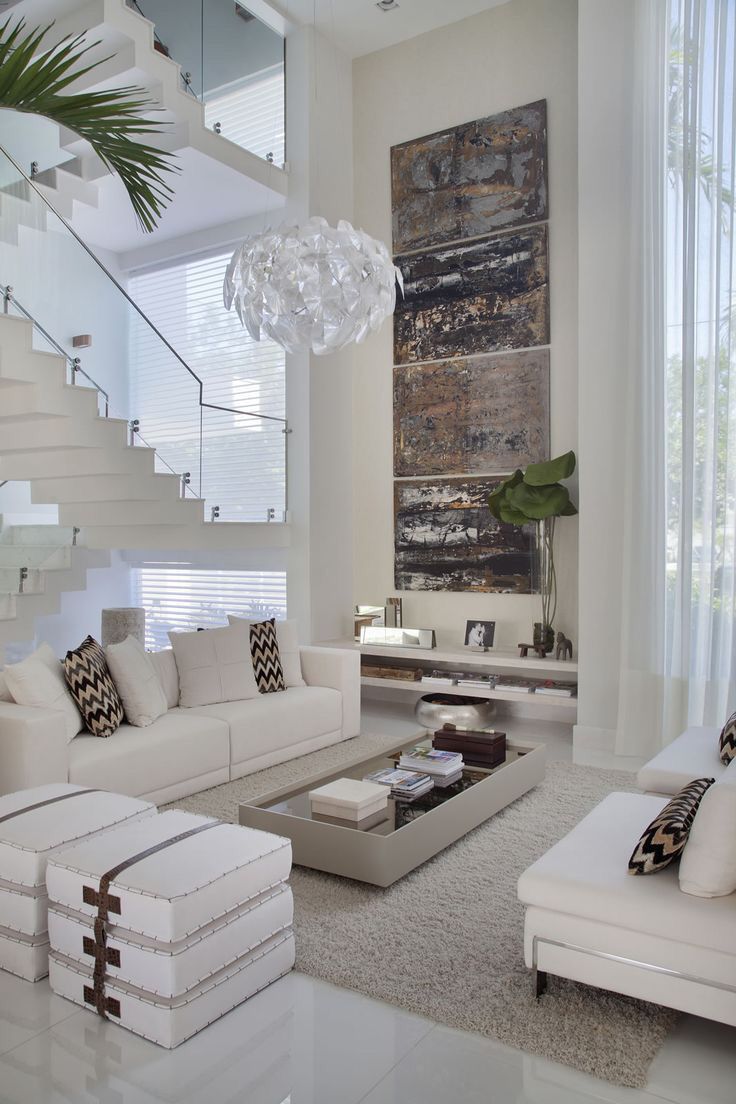} 
	\caption{Modern}
	\label{fig:interiordesign_l}
\end{subfigure}
\begin{subfigure}[b]{0.15\textwidth}
	\centering
	\includegraphics[height=0.5\textwidth]{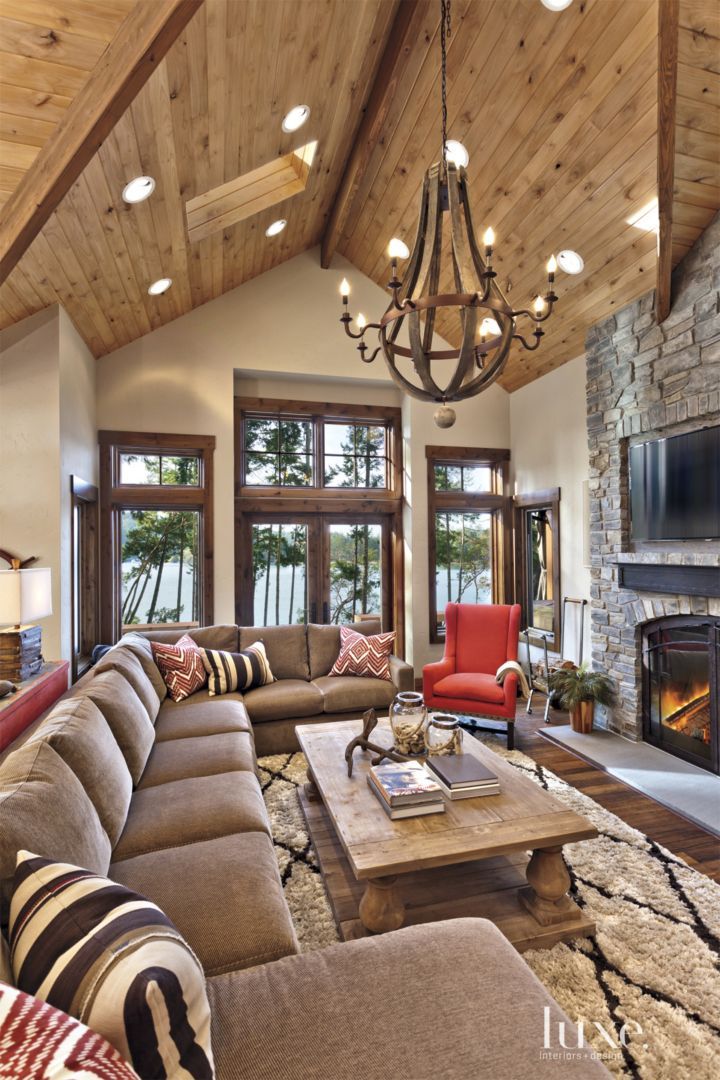} 
	\caption{Traditional}
	\label{fig:interiordesign_r}
\end{subfigure}
\quad
\begin{subfigure}[b]{0.15\textwidth}
	\centering
	\includegraphics[height=0.5\textwidth]{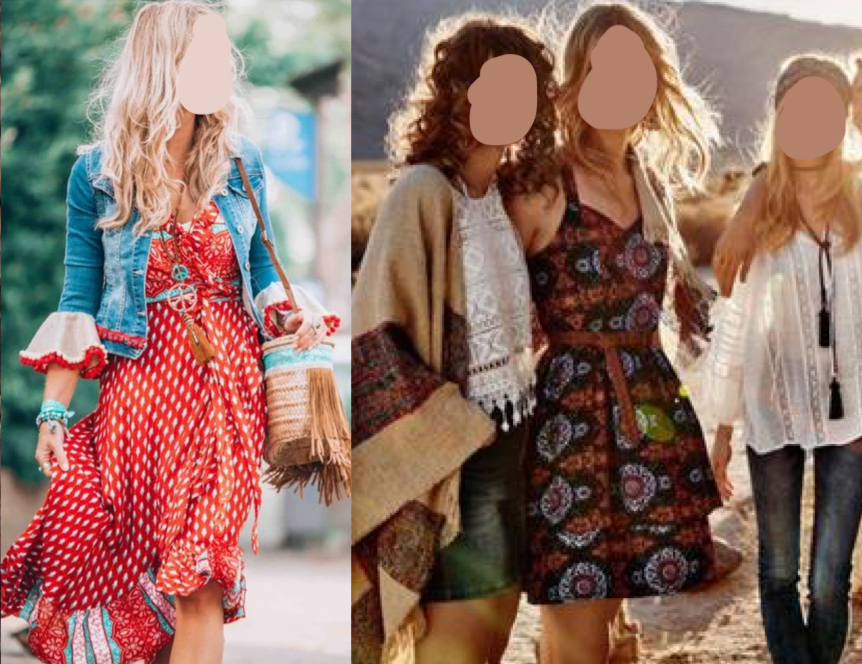} 
	\caption{Bohemian Chic}
	\label{fig:fashion_l}
\end{subfigure}
\begin{subfigure}[b]{0.15\textwidth}
	\centering
	\includegraphics[height=0.5\textwidth]{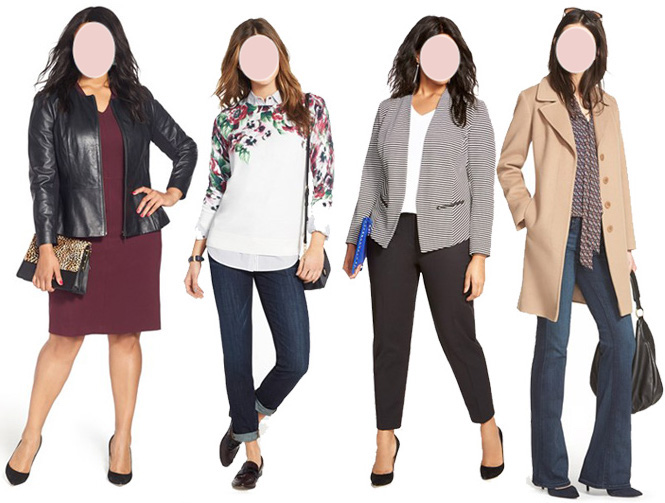} 
	\caption{Business casual}
	\label{fig:fashion_r}
\end{subfigure}
\caption{Example images shown to subjects when asking for their preferences in paintings: Cubism vs. Renaissance, interior design: Modern vs. Traditional, and fashion: Bohemian Chic vs. Business Casual. Other than showing them as examples to remind subjects of what these categories mean, these images were not used in this work in any way.}
    \label{fig:life}
\end{figure*}

The generative artist designed this generative process, chose the colors in each palette, the probability $P$ with which a new shape starts after each curve, the range $R$ that determines the amount of noise added to a mid point, and the probability $\nu$ of adding noise to each of the 4 directions to form the control point of the quadratic Bezier curve (when applicable). The machine picks the random end points and noise added to the mid point to form the control point of the quadratic Bezier curve (when applicable). The color palette, number of curves $N$ in the piece, thickness of curves $T$ and whether the curves should be a straight line or a quadratic Bezier curve are free parameters. These parameters are provided as options on an interactive tool as seen in Figure~\ref{fig:interactive_strokes}. The random seed is kept fixed when a user is changing parameters on the interface so that the only influence changing the piece is input from the user. The user can click on the ``Generate'' button to change the random seed that determines the machine's influence.  A video demonstrating the interface is provided here \url{https://youtu.be/YzfzjK8NNMg}.


As options, the tool provides 6 color palettes, 11 densities which determine the number of curves ($N = 2^\text{density}$), 15 line thicknesses, and a binary option of curved or straight lines. In our study, which serves as a proof of concept, we restrict the number of palettes to 3, densities to 3, line thicknesses to 4, and retain the straight vs. curved lines option.\footnote{Conceptually, our approach is not restricted to a few parameters, because it learns correlations in preferences offline across a population of users, rather than online for an individual user.} Specifically, we start with a ``default'' configuration for each of these options (Figure~\ref{fig:default_vs_alternatives}a), and generate 8 versions of the piece by changing one property at a time (Figure~\ref{fig:default_vs_alternatives}b-i): switching to one of the other two palettes (gray in Figure~\ref{fig:default_vs_alternatives}c and bright in Figure~\ref{fig:default_vs_alternatives}g), changing the density to one of the other two values (sparser in Figure~\ref{fig:default_vs_alternatives}d and sparsest in Figure~\ref{fig:default_vs_alternatives}e), changing the line thickness to one of the other three values (thin in Figure~\ref{fig:default_vs_alternatives}i, thicker in Figure~\ref{fig:default_vs_alternatives}b, and thickest in Figure~\ref{fig:default_vs_alternatives}h), and switching to straight lines (Figure~\ref{fig:default_vs_alternatives}f). 

\section{Collecting Preferences}

To answer our research question, we collect two classes of preferences from subjects. One is in the context of the Strokes interactive generative art described above. The other is about other walks of life. Each preference is posed as a two-way forced choice question.

For Strokes, we generate 8 pairs of comparisons: default in Figure~\ref{fig:default_vs_alternatives}a vs. each of the 8 edited versions in Figure~\ref{fig:default_vs_alternatives}b-i. Both pieces in each pair are generated with the same random seed so that there is only one cause of variation between the two pieces, but different seeds are used across pairs to ensure that the preferences we collect are generic across seeds. We randomly order the default and the edited version within a pair. The 8 pairs are also randomly ordered. We generate a total of 3 sets of these comparisons with different random seeds. This gives us 24 two-way forced choice pairs in the context of interactive generative art. For each pair, subjects were asked ``Which visual pattern appeals to you more?''

For other walks of life, we ask subjects for 12 preferences.
\begin{packed_itemize}
\item Do you reflect on a regular basis (e.g., write in a journal)? Yes/No
\item Which do you prefer? Milk vs. dark chocolate.
\item Which do you prefer? Wine vs. beer.
\item Which do you prefer? Country vs. rock music.
\item Do you have any exposure to design principles? Yes/No.
\item Are you artistically inclined? Yes/No
\item Which gender do you associate with more? Male/female.
\item What personality type do you associate with more? Introvert/extrovert.
\item Which do you prefer? Sweet vs. savory food.
\item Which of these styles of painting appeals to you more? Cubism	 vs. Renaissance (Figures~\ref{fig:painting_l}, \ref{fig:painting_r} were shown).
\item If you could setup your home however you liked, which of these styles would you go with? Modern vs. traditional
(Figures~\ref{fig:interiordesign_l}, \ref{fig:interiordesign_r} were shown as examples).
\item Irrespective of your gender, which of these fashion styles do you relate to more? Bohemian Chic vs. business casual
 (Figures~\ref{fig:fashion_l}, \ref{fig:fashion_r} were shown as examples).
 \end{packed_itemize}

These 12 combined with the 24 generative art preferences described earlier gives us a total of 36 two-way forced choice questions. We collected these 36 preferences from 311 subjects on Amazon Mechanical Turk. 

\section{Analyzing Preferences}

We begin by analyzing the preferences. We first ask how consistent subjects are in their own preferences in generative art. Recall that each subject was shown all 8 generative art comparisons for 3 different seeds. If preferences change significantly across random seeds, there is limited scope to predict them. But if the choices are consistent across seeds, there may be scope to predict them automatically.

\subsection{Are people consistent?}

For the 3 sets of responses for each of the 311 subjects and each of the 8 comparisons (i.e., across a total of 2488 sets of 3 responses), we find that on average 2.7 responses are the same. 2.7 out of 30 is 90\%. Random guessing would result in 75\% of the responses being the same. 

Another way of looking at inter-human agreement is to check how accurately a person's response to the 3rd random seed can be predicted if one assumes that their response will be the same as their response to the other 2 random seeds. We find this accuracy to be 81\%. Recall that each pair contains the default configuration (Figure~\ref{fig:default_vs_alternatives}a) and one of the eight alternative versions (Figure~\ref{fig:default_vs_alternatives}b-i). We find that across subjects, the default configuration  is preferred 62\% of the time. This is not surprising because the default configuration has been set by the generative artist as a generically good configuration. We use this prior preference for the default configuration to break ties. That is, in cases where the response to the first two random seeds was different (one picking the default and the other picking the alternative version), we assume that for the 3rd seed the subject picked the default version. Note that always assuming prior -- that is, always predicting that the user prefers the default options -- would result in a prediction accuracy of 62\% -- significantly lower than 81\% reported above.

This suggests that people prefer alternative configurations frequently \emph{and} that they are consistent in this preference across random seeds. That is, there is scope for predicting the personal preferences of a user automatically.

\subsection{What do people like?}

Which configurations do subjects tend to like better? Figure~\ref{fig:default_vs_alternatives} shows the \% of times each alternative is preferred over the default. As discussed above, the default is most preferred (62\% of the time), but there is a heavy tail suggesting that the other alternatives are preferred frequently. We find that on average across subjects, 3 out of the 8 alternatives are preferred over the default.  As for preferences in other walks of life, Table~\ref{table:preferences} shows the \% of times the various options were preferred. Most preferences are not significantly skewed in one direction over the other, and may have useful information to predict other preferences.

\begin{table}[h]
\caption{Subjects' preferences in various walks of life.}
\begin{tabular}{@{\extracolsep{\fill}}p{6.5cm}c@{\extracolsep{\fill}}}
\midrule
\multicolumn{2}{c}{What do subjects like?} \\
Milk over dark chocolate  &  60\% \\
Sweet over savory food & 34\% \\
Wine over beer &  53\% \\
Country over rock music & 25\% \\
Cubism over Renaissance & 43\% \\
Modern over traditional interior & 34\% \\
Bohemian over business casual & 43\% \\
\midrule
\multicolumn{2}{c}{What are subjects like?} \\
Male & 39\% \\
Reflect regularly & 40\% \\
Exposure to design principles & 26\% \\
Artistically inclined & 48\% \\
Introvert & 74\% \\
\midrule
\end{tabular}
\label{table:preferences}
\end{table}

\subsection{Which preferences are related?}

Next, we analyze which preferences are most related to each other. For instance, if we wanted to know whether a person will like the gray palette over the default in Strokes, is it best to use information about whether they liked straight over curved lines in Strokes (perhaps  those who prefer straight lines prefer the gray palette)? Or whether they like Bohemian Chic fashion over Business Casual? That is, for a target preference of interest, which other preference is the best source such than if the source preference is directly used to predict the target preference, prediction accuracy is the highest across all potential source preferences. 

To evaluate this, for each of the 36 target preferences, we search over the remaining 35 preferences as a potential source preference and compute the prediction accuracy. We report the best source for some target preferences in Table~\ref{table:best_source}. We can compare this prediction accuracy (Acc) to the prior accuracy (Prior) -- the accuracy we would get if for every target preference we predicted whichever choice is the most common. In addition to reporting the best source for some target preferences, in Table~\ref{table:best_source} we also report some source preferences that are not necessarily the best, but are still more accurate than this baseline and we thought showcase interesting correlations in our data.

\begin{table}[h]
    \caption{Analysis of which (source) preferences are good indicators to predict other (target) preferences.}
\begin{tabular}{@{\extracolsep{\fill}}p{2.9cm}p{2.9cm}cc@{\extracolsep{\fill}}}
\midrule
Target & Source & Acc & Prior\\
\multicolumn{4}{c}{Best source to predict a target preference} \\
Reflect regularly  & Design exposure & 61\% & 59\% \\
Wine over beer & Female & 65\% & 53\% \\
Artistic & Design exposure & 66\% & 51\% \\
Male & Beer over wine & 65\% & 60\% \\
Grey palette & Straight lines & 61\% & 59\% \\
Sparsest pattern & Sparser pattern & 73\% & 61\% \\
Sparser pattern & Sparsest pattern & 73\% & 60\% \\
Thicker lines & Thickest lines & 77\% & 59\% \\
Thickest lines & Thicker lines & 77\% & 66\% \\
\multicolumn{4}{c}{Interesting source to predict a target preference} \\
Artistic & Bohemian$>$business & 56\% & 51\% \\
Thicker lines & Bright palette & 61\% & 58\% \\
Sparser pattern & Thin lines & 62\% & 60\% \\
Artistic & Sparser pattern & 53\% & 51\% \\
Not artistic & Milk over dark choc. & 53\% & 51\% \\
\midrule
\end{tabular}
\label{table:best_source}
\end{table}

We see that gender is the best predictor for whether someone prefers wine or beer. Whether someone prefers the gray palette over the default is the best predictor for whether they prefer straight lines over the curved lines in Strokes. Whether someone prefers the bright palette is a good indicator of whether they prefer thicker lines in Strokes. Unsurprisingly, whether someone prefers the sparser pattern is best predicted by whether they preferred the sparsest pattern. Overall, we see promising indications that there are correlations between the different preferences, and there is potential to predict a subset of preferences from other preferences. To build these predictors, we train a variety of machine learning models, which we describe next.

\section{Predicting Preferences}

In order to predict a subset of preferences from other preferences, we train machine learning models. Our setup is the following: We create three groups of preferences -- $\mathcal{A}$: The 8 art (Strokes) preferences (recall that we have 3 responses from each subject for these 8 preferences), $\mathcal{L}$: the 12 preferences from other walks of life, and $\mathcal{U}$: all 20 preferences.

Each group can be used as features $\mathcal{F}$ to predict each of the preferences in the other group (target $\mathcal{T}$). A group can also be used as features to predict a preference from the same group by excluding that target preference from the input features. This results in a total of 9 machine learning problems $(\mathcal{F},\mathcal{T}) \in \{\mathcal{A},\mathcal{L},\mathcal{U}\} \times \{\mathcal{A},\mathcal{L},\mathcal{U}\}$ where $\times$ denotes the cartesian product. These include (as non-exhaustive examples):

\begin{packed_itemize} 
\item Predicting art preferences from other life preferences (i.e., features: $\mathcal{A}$, target: each preference from $\mathcal{L}$, metric: average accuracy across all preferences in $\mathcal{L}$).
\item Predicting other life preferences from art preferences (i.e., features: $\mathcal{L}$, target: each preference from $\mathcal{A}$, metric: average accuracy across all preferences in $\mathcal{A}$).
\item Predicting an art preference from other art preferences.
\end{packed_itemize}

We train and test our machine learning models via leave-one-out cross validation. We train on preferences from 310 of 311 subjects, and test on the remaining subject. This is done 311 times (rotating which subject we test on). We report average accuracies across the 311 tests. Recall that we collected preferences from each subject for the 8 Strokes preferences across 3 seeds. We ensure that all 3 preferences from a subject are either in test or train, and not split across. 

To normalize for the fact that different preferences have a different prior, we report class normalized accuracies. Class normalized accuracy is the average of the accuracies of predicting a subject’s preference in both directions. So if we predict that a subject always prefers the default Strokes piece (say), we will have 100\% accuracy for class `default' but will have missed all instances when the subject preferred the alternative, leading to 0\% accuracy for class `alternative'. This results in a normalized class accuracy of 50\% (average of 0\% and 100\%). That is, class normalized accuracy of prior is 50\% and is the baseline we compare our models to. 

We experiment with the following approaches: nearest neighbor, logistic regression, linear Support Vector Machines (SVMs), polynomial SVMs, Radial Basis Function (RBF) SVMs, neural networks, and decision trees, as well as a matrix completion approach. Hyper parameters of each of these models are picked via cross validation. 

\begin{figure}[t]
    \centering
    \includegraphics[width=0.7\columnwidth]{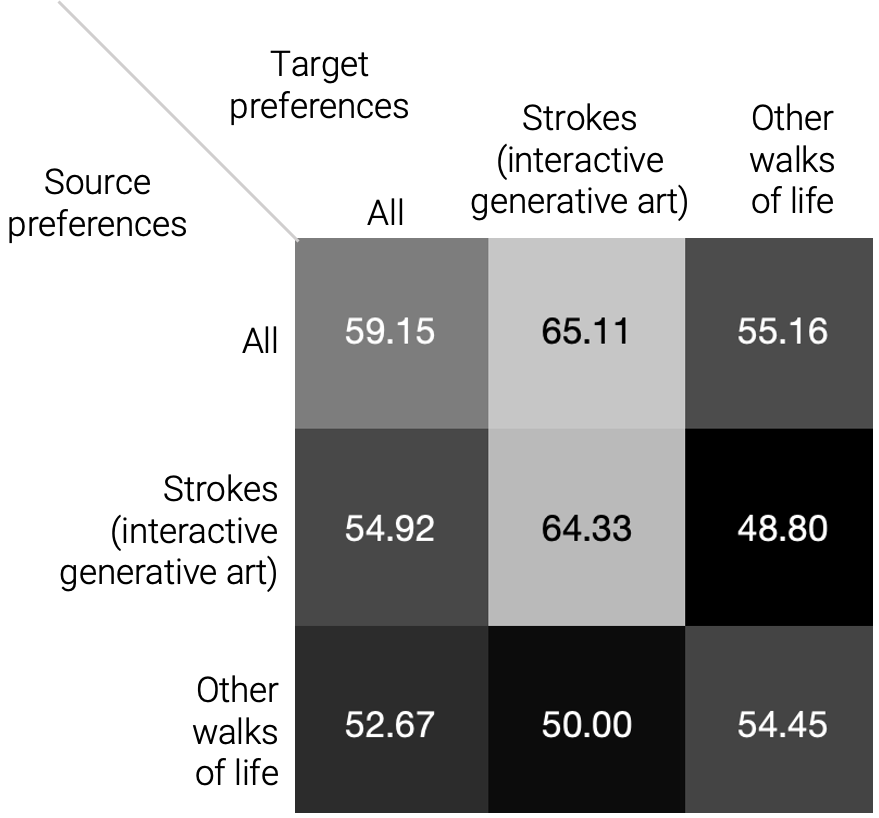}
    \caption{Accuracies across all 9 settings corresponding to choices of source preferences (features $\mathcal{F}$) and target preferences (prediction targets $\mathcal{T}$) using a linear SVM.}
    \label{fig:accuracy_matrix}
\end{figure}

The linear SVM performs the best. In Figure~\ref{fig:accuracy_matrix} we show its accuracy on all 9 configurations of the machine learning setup described earlier. Recall that performance of a prior baseline under our normalized accuracy metric is 50\%. We see  that our models are unable to predict interactive generative art (Strokes) preferences based on other life preferences and vice versa. However Strokes preferences predict other Strokes preferences well. Knowing other preferences in life further (incrementally) helps predict Strokes preferences. We provide further analysis of our machine learning models for these two settings: using interactive generative art (Strokes) preferences to predict other Strokes preferences, and using all preferences (Strokes + other preferences in life) to predict a held out Strokes preference.

\begin{figure}[t]
    \centering
    \includegraphics[width=\columnwidth]{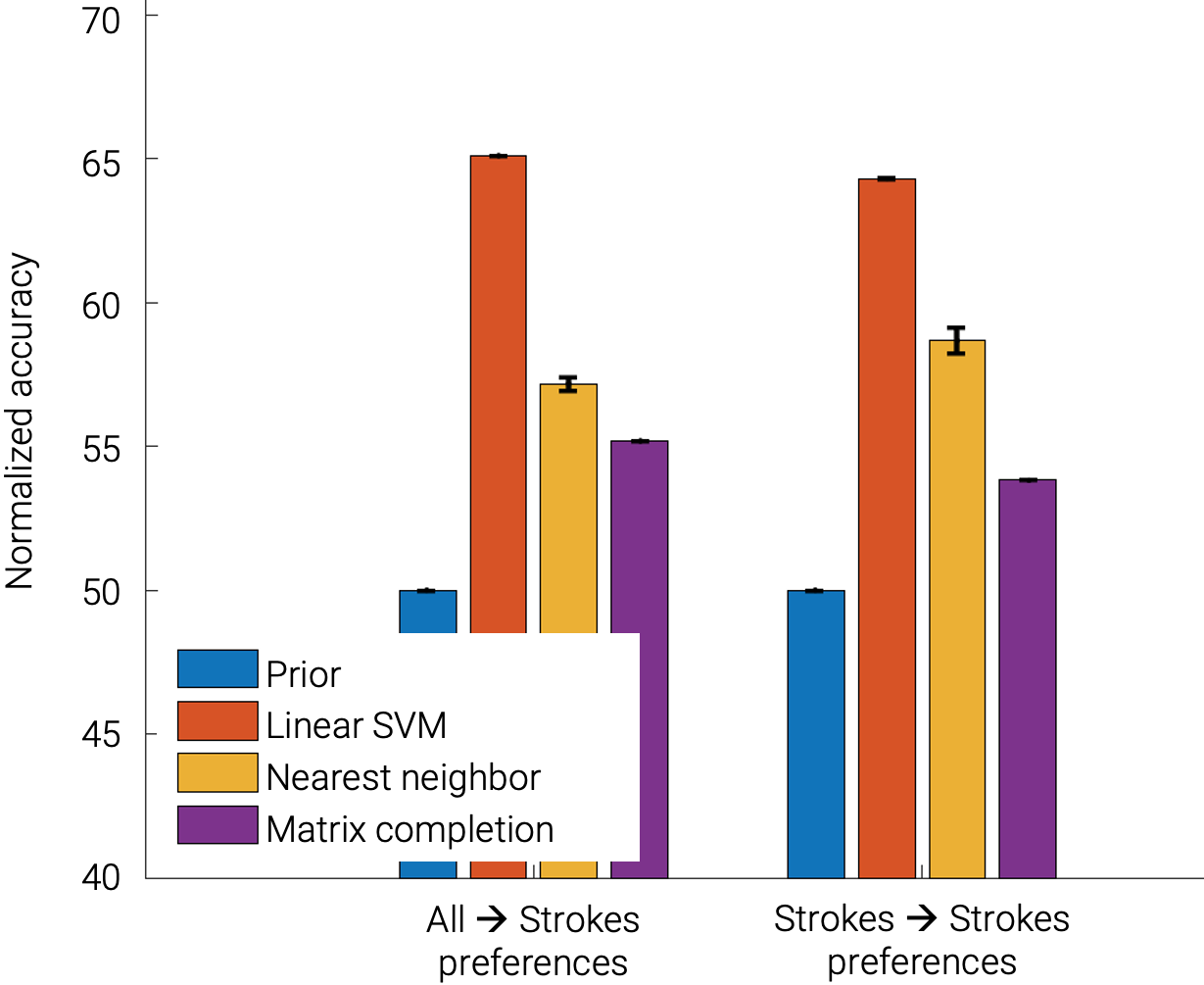}
    \caption{Comparison of machine learning models.}
    \label{fig:classifiers_comparison}
\end{figure}

While linear SVMs performed the best, nearest neighbor and matrix completion accuracies are informative points of comparison. The nearest neighbor approach assumes that a test subject's preference is the same as the training subject who has the most number of other preferences is common with the test subject. Matrix completion approaches are commonly used in recommendation systems to predict user preferences. In Figure~\ref{fig:classifiers_comparison} we show a comparison of these classifiers along with the prior baseline.  As an additional point of reference, if we use the best single source preference to predict each target preference (similar to the analysis from Table~\ref{table:best_source}), we get a prediction (class normalized) accuracy of 60\%. This is lower than the accuracy of the linear classifier, demonstrating the benefit of using information from multiple preferences to make the prediction. 

The linear SVM, matrix completion and baseline prior approaches are deterministic. The nearest neighbor approach often finds multiple nearest neighbors (all equidistant from the test instance). We break ties randomly. To assess statistical significance, Figure~\ref{fig:classifiers_comparison} shows variation due to this stochasticity via a 95\% confidence interval. As another measure of statistical significance, particularly for the deterministic approaches, we use 1,000 bootstrap samples and find the 95\% confidence interval to consistently be $\pm$$\sim$0.1\%.

\subsection{Interpreting Predictions}

\begin{figure}[t]
    \centering
    \includegraphics[width=\columnwidth]{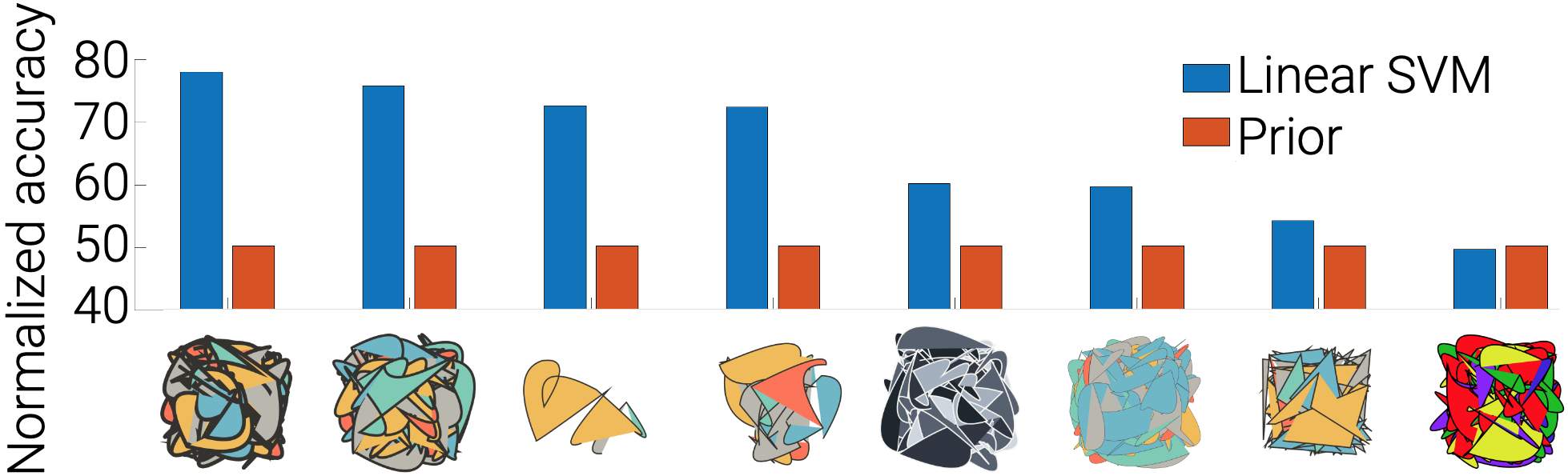}
    \caption{Predictability of preferences in interactive generative art (Strokes) from all other preferences of a user.}
    \label{fig:easiest_to_predict}
\end{figure}

\begin{figure*}[t]
    \centering
    \includegraphics[width=0.95\textwidth]{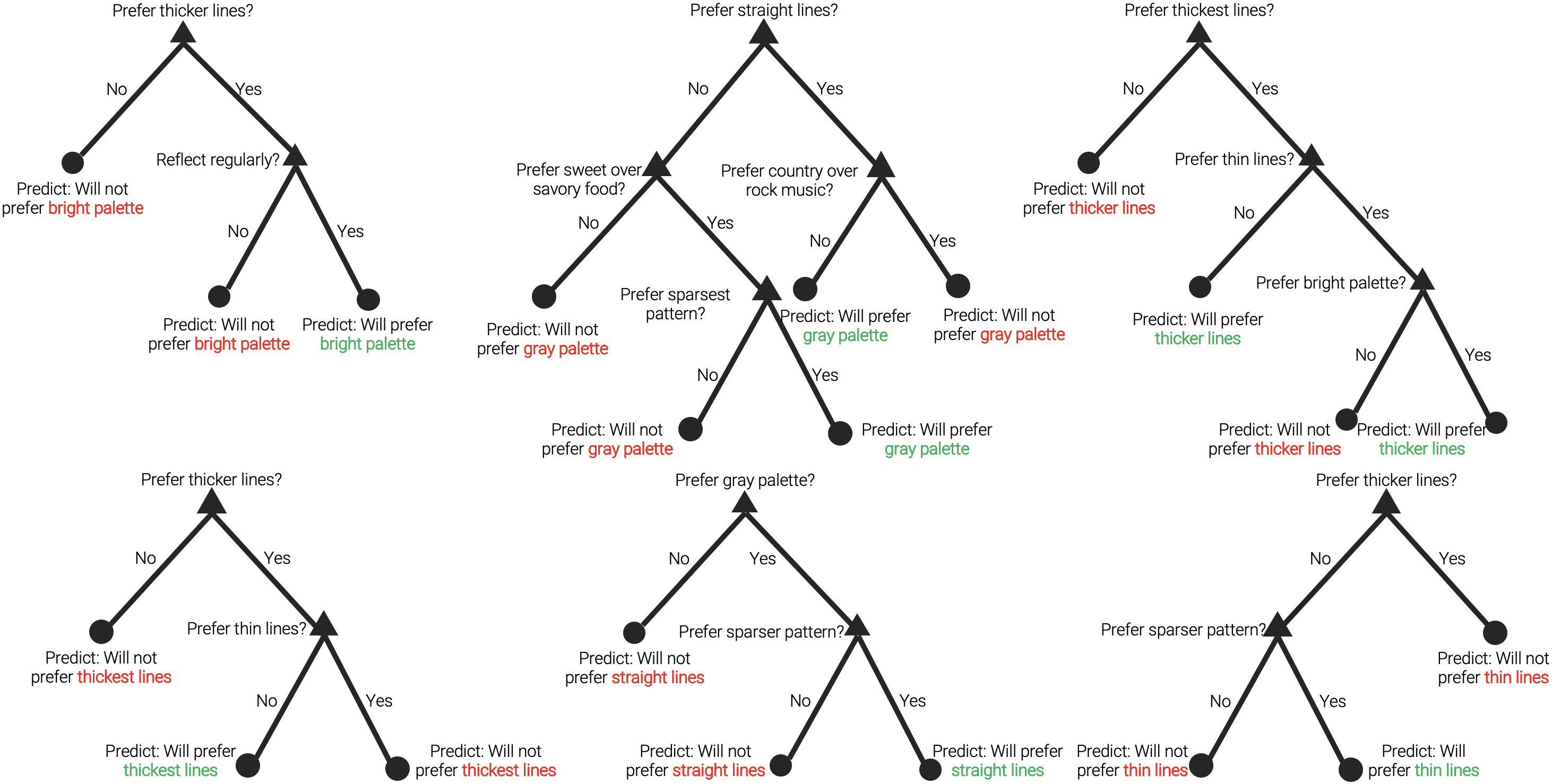}
    \caption{Decision trees to interpret the multi-dimensional correlations between different preferences in our data.}
    \label{fig:decision_trees}
\end{figure*}

\begin{figure}[t!]
    \centering
    \includegraphics[width=0.95\columnwidth]{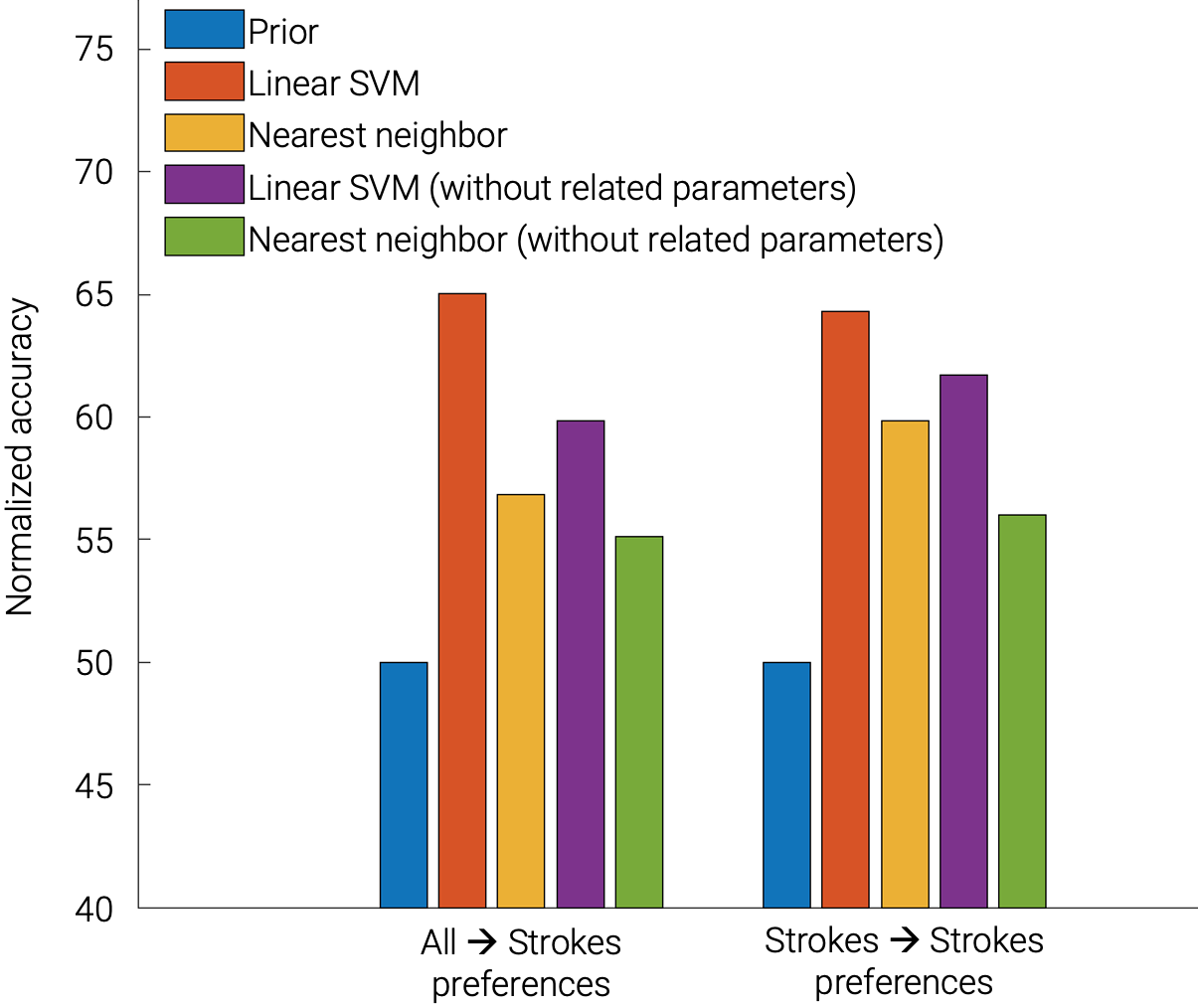}
    \caption{Accuracies at predicting Strokes preferences before (left) and after (right) removing Strokes preferences that are linearly related (e.g., thicker vs. thickest lines).}
    \label{fig:no_cross_params_classifiers}
\end{figure}

We now analyze the machine learning models to get a better insight into the patterns in the data. We start by analyzing which of the Strokes preferences are easiest to predict i.e., are most related to all other preferences of a user (within Strokes and in other walks of life). See Figure~\ref{fig:easiest_to_predict}. We see that it is easiest to predict whether someone will like the thickest setting of the line width or not. On the other hand, it is hardest to predict whether someone will like the bright color palette or not (we cannot guess any better than the prior).

Next, we seek interpretable ``rules'' that explain how different preferences are related to each other. For this we turn to decision trees as interpretable machine learning models. We train a decision tree on all preferences except a target Strokes preference that we wish to predict. To aid in interpretability, we restrict the depth of the decision tree to be 3 levels or less. Figure~\ref{fig:decision_trees} show several instances of rules that we recover for different target preferences.

We see that if someone prefers the gray palette, and if they prefer the sparser pattern, they are likely to prefer straight lines. Otherwise (either they prefer the default palette over the gray one or over the sparser pattern), they are likely to prefer the curved lines over the straight lines. If someone prefers straight lines and country music, they are likely to prefer the default palette over the gray palette. Note that these patterns give interesting insights into the data we collected and inspire thought about how one might predict a user's preferences in interactive generative art. They are are not meant to encourage generalization or stereotypes.

One thing that stands out in Figure~\ref{fig:decision_trees} is that whether someone will prefer thicker lines can be predicted by checking if they like the thickest or thin lines. Same for sparse vs. dense patterns. These preferences along a linear ordering are obviously (and hence somewhat uninterestingly) related (if someone does not like sparse patterns, it is unlikely they will like even sparser patterns). 

To verify that the predictive power of our machine learning models is not relying primarily on these uninteresting correlations, we reduce our set of Strokes preferences down to 5. We remove the sparsest (Figure~\ref{fig:default_vs_alternatives}e), thickest lines (Figure~\ref{fig:default_vs_alternatives}h) and thin lines (Figure~\ref{fig:default_vs_alternatives}i) alternatives because those were the least preferred alternatives along the thickness and density parameters. The remaining 5 alternatives are all orthogonal to each other (with no linear ordering). We retrain our machine learning models. Their performance is shown in Figure~\ref{fig:no_cross_params_classifiers}. 
We see that while model accuracies go down a little after removing these linear dependencies (e.g., accuracy of a linear SVM at predicting Strokes preferences from other Strokes preferences goes from 64\% to 62\%), it continues to be significantly better than the prior baseline (50\%). This suggests that we can indeed predict meaningful (non-obvious) dependences between a user's preferences when interactively creating generative art (Strokes).

\begin{figure}[t]
    \centering
    \includegraphics[width=0.9\columnwidth]{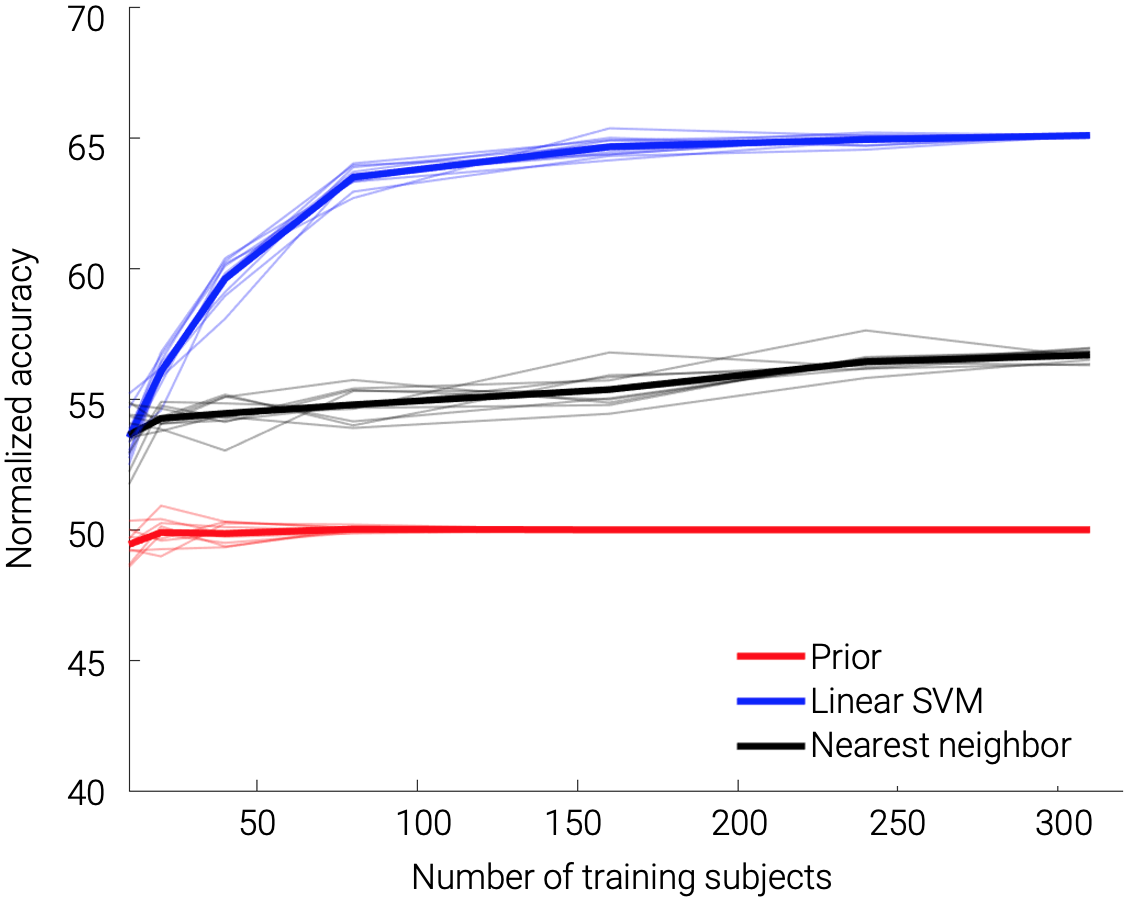}
    \caption{Accuracy with increasing training data.}
    \label{fig:amount_of_data}
\end{figure}

\vspace{5pt}
\noindent \textbf{More data?}
Finally, we ask the question: would more training data improve the performance of these models? Figure~\ref{fig:amount_of_data} shows the performance of our models at predicting the Strokes preferences from all (Strokes + other walks of life) but the target preference, with increasing amount of data. For a fixed amount (x-axis), we show accuracies when training on 10 random subsets of data of that size. We find that performance of the linear model has saturated. Performance of a nearest neighbor model, as expected, continues to increase with more training data. The more subjects we collect preferences from, the more likely it is that a subject in our training set will closely mimic a new test subject.

\section{Conclusion and Future Work}

There are numerous avenues for future work. In this work the interactive generative art preferences involved studying just one generative art form, always comparing an alternative configuration with a default configuration. Expanding the study to include comparisons between two alternative configurations, other generative art forms, and predicting preferences across generative art forms is future work. 

Future work also includes translating the machine learning models we trained in this paper into a smarter interactive generative art tool. This leads to more complex machine learning problems of modeling the sequence of interactions (as opposed to one-shot preferences we modeled in this paper), and determining if the machine learning models should be used to eliminate part of the parameter space or promote a part of the parameter space. This involves focussing on either precision or recall of the models.

To summarize, while we have not yet found evidence for it, given the narrow scope of our study, there may still be potential in the use of interactive generative art creation as an engaging and creative ``personality test''. In fact, it may be possible to design generative art that explicitly optimizes for correlation with personality traits or preferences in other walks of life. We \emph{do} find evidence that preferences of a user creating art using an interactive generative art tool are predictable from choices they make. This opens up opportunities for smart casual creators that make it easier for a lay person to create a piece they are personally excited about! 

\section{Acknowledgments}


The Strokes interactive generative art tool was made in p5.js. ``p5.js is a JavaScript library for creative coding, with a focus on making coding accessible and inclusive for artists, designers, educators, beginners, and anyone else!'' Thank you Larry Zitnick for helpful discussions.

\small
\bibliographystyle{iccc}
\bibliography{iccc}

\end{document}